\documentclass[letterpaper, 10 pt, conference]{ieeeconf} 
\usepackage{times}
\usepackage{amsmath}
\usepackage{xcolor}
\usepackage{mathtools}
\usepackage{booktabs}
\usepackage{amssymb}
\usepackage{multicol}
\usepackage[skip=2pt,font=small]{caption}
\usepackage{subcaption}
\usepackage{bm}
\usepackage[backend=biber,style=ieee,mincitenames=1,maxnames=20,maxcitenames=1,natbib=true]{biblatex}
\IEEEoverridecommandlockouts                              

\title{\LARGE \bf
Learning Deep Dynamical Systems using Stable Neural ODEs
}

\author{Andreas Sochopoulos$^{1, 2}$ Michael Gienger$^{2}$ and Sethu Vijayakumar$^{1, 3}$
\thanks{$^{1}$ School of Informatics, The University of Edinburgh, Edinburgh, UK}%
\thanks{$^{2}$ Honda Research Institute Europe GmbH, Offenbach am Main, Germany}%
\thanks{$^{3}$ The Alan Turing Institute, London, U.K.}
}

\begin{document}

\maketitle
\thispagestyle{empty}
\pagestyle{empty}

\begin{abstract}
Learning complex trajectories from demonstrations in robotic tasks has been effectively addressed through the utilization of Dynamical Systems (DS). State-of-the-art DS learning methods ensure stability of the generated trajectories; however, they have three shortcomings: a) the DS is assumed to have a single attractor, which limits the diversity of tasks it can achieve, b) state derivative information is assumed to be available in the learning process and c) the state of the DS is assumed to be measurable at inference time. We propose a class of provably stable latent DS with possibly multiple attractors, that inherit the training methods of Neural Ordinary Differential Equations, thus, dropping the dependency on state derivative information. A diffeomorphic mapping for the output and a loss that captures time-invariant trajectory similarity are proposed. We validate the efficacy of our approach through experiments conducted on a public dataset of handwritten shapes and within a simulated object manipulation task.
\end{abstract}

\section{Introduction}
\label{sec:introduction}
Learning from Demonstrations (LfD) is a methodology that allows robots to learn complex behaviours by closely observing and emulating human or expert demonstrations \cite{calinon2018learning}, that would otherwise be too cumbersome to compute or hand code. By leveraging this method, robots can bridge the gap between human expertise and machine capabilities, facilitating rapid skill acquisition and adaptation. Replicating human motions can also greatly benefit the collaboration capabilities of the robot in Human Robot Interaction (HRI) tasks and act in more predictable ways for humans. Moreover, the flexibility of the framework, allows the robot to learn how humans move in order to use it as a predictive model rather than replicate their movement. Creating surrogate models for humans has shown good potential in HRI \cite{fishman2020collaborative} and exoskeleton assistance adaptation tasks \cite{sochopoulos2023human}.  
\begin{figure}[t!]
    \centering
    \includegraphics[width=\linewidth]{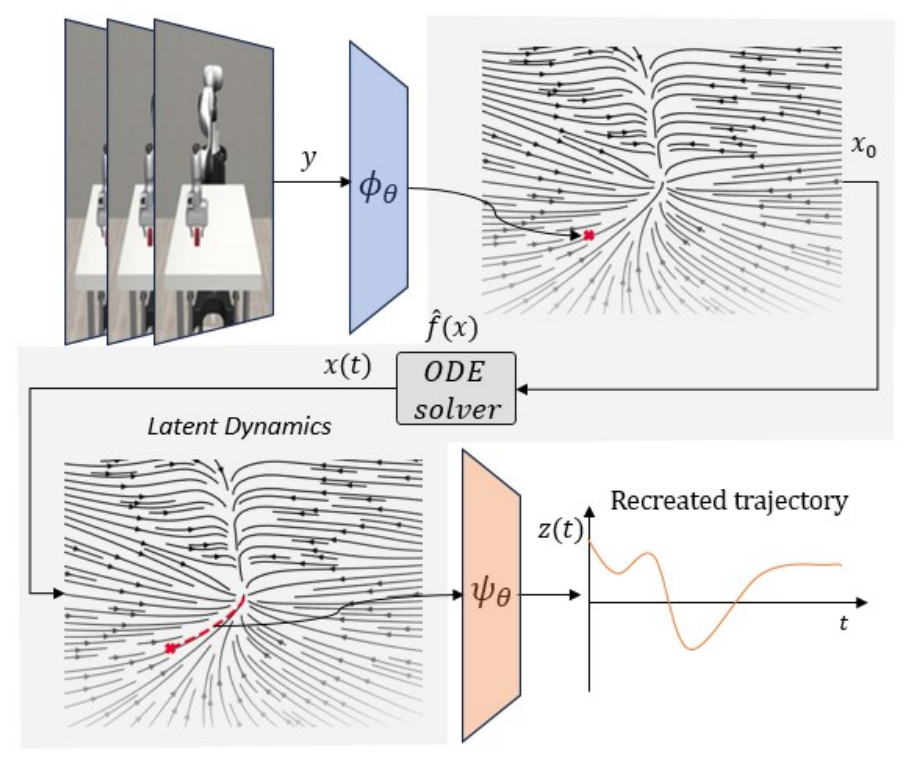}
    \caption{End-to-End learning framework for Dynamical Systems using stable Neural ODEs}
    \label{fig:figure1}
\end{figure}

Human motions often exhibit complex patterns, smoothness, and stability. However, encoding these constraints on learnt models is non-trivial. Dynamical Systems (DS) provide a versatile framework for encoding such trajectories with intrinsic structures that, if properly constructed, ensure stability, smoothness, and the ability to capture intricate spatio-temporal dependencies. 

Stability in learning methods is typically addressed in terms of globally stable attractors, limiting the system to single-goal-driven motions. Although most tasks can be split up in multiple goal-driven motions \cite{schaal1999imitation}, identifying the hierarchy of subtasks and their hyperparameters is non-trivial. DS with multiple attractors \cite{manschitz2018mixture} and DS whose trajectories are constrained to move in a subset of the phase space can represent more complex behaviours and increase the variety of tasks a DS can achieve. Moreover, many existing approaches in LfD assume that the state whose dynamics are being captured is readily available for measurement. However, in many robotic tasks, this assumption does not hold true due to either a lack of sensing capabilities or a mismatch between sensing and state modalities. For example, in tasks where only vision is available as a sensing modality, extracting the end effector pose from images adds complexity and can compromise accuracy. Additionally, most methods rely on state derivative information \cite{khansari2011learning, khansari2014learning, rana2020euclideanizing}, such as position, velocity, and acceleration, treating the learning process as a regression problem. This requirement adds complexity to the demonstration acquisition process, necessitating additional hardware and data processing. The quality of the data is crucial as well, as the system learns to reproduce demonstrated trajectories with their timing intact. Speeding up demonstrations can lead to a completely different learned model, making reproducibility challenging and complicating the acquisition process further.

\subsection{Related Works}
Dynamic Movement Primitives (DMPs) \cite{ijspeert2013dynamical} were among the first approaches to utilize DS for learning robot motions. The DS learned through DMPs typically consist of a stable linear component and a nonlinear forcing term composed of Radial Basis Functions (RBFs) with weights adapted to closely match demonstrated trajectories. Subsequently, the first approach to approximate the entire DS with stability guarantees was SEDS \cite{khansari2011learning}, which employed Gaussian Processes (GPs) as function approximators for the dynamics. Lyapunov stability is ensured for a single goal state by imposing constraints on the weights of the GPs. However, the linear nature of DMPs and the constraints of SEDS limit their representational capabilities, allowing only for the learning of simple motions evolving in low-dimensional phase spaces.

The trade-off between accuracy and stability has been addressed in later methods \cite{kolter2019learning, khansari2014learning}, which assume an unconstrained and potentially unstable part of the dynamics alongside a corrective term that ensures global stability, in a Lyapunov sense, of a designated attractor. Additionally, other methods have achieved increased accuracy by employing diffeomorphisms \cite{neumann2015learning, rana2020euclideanizing, zhang2022learning}. Specifically, these methods construct a simple and provably stable DS in a latent space, which is then mapped through parameterized diffeomorphisms in the output space. This learning paradigm updates only the weights of the parameterized diffeomorphism rather than the state mapping of the DS. The stability of the output space system is guaranteed since it is diffeomorphic to the latent space system which is stable by design. Lastly, some approaches have capitalized on the non-Euclidean nature of the data encountered in robot applications and have introduced DS that evolve in manifolds \cite{rana2020euclideanizing, zhang2022learning}.

\begin{table*}[ht!]
    \centering
    \caption{Comparison of DS Learning Methods based on their capacity to learn different types of DS. The first column shows the dependence of the learning methods on state derivative information. The second and third column indicate the ability of the model to learn DS with multiple attractors and DS that live in latent spaces respectively. Lastly stability is referred to the Lyapunov stability of the learnt DS. *: Although multiple attractors can be potentially handled by \cite{rana2020euclideanizing}, no method to achieve this is given.}
    \begin{tabular}{lccccc}
        \toprule
        & State Derivative Free & Multiple attractors & Multiple Input Modalities & Stable \\
        \midrule
        CLFDM \cite{khansari2014learning} & --& -- & -- & \checkmark \\
        Deep Dynamics \cite{kolter2019learning} & \checkmark& --& \checkmark& \checkmark\\
        Euclideanizing Flows \cite{rana2020euclideanizing}& -- & \ \checkmark$^*$ & -- & \checkmark \\
        \midrule
        \textbf{StableNODE} (Ours) & \checkmark& \checkmark& \checkmark & \checkmark \\
        \bottomrule
    \end{tabular}
    \label{table}
\end{table*}

DS offer spatiotemporal representations that have been utilized in machine learning as well. Neural ODEs \cite{chen2018neural} have emerged as a method to encode the hidden layers of Residual Networks as a continuous time ODE. They are a class of architectures that combine neural networks with dynamical systems and thus naturally fit into the problem of learning DS for robot and control applications \cite{miao2023learning, bottcher2022ai, bottcher2022near}. 
\subsection{Contributions}
In this paper, we use Neural ODEs to represent and train DS with deep state mappings. Since they lack intrinsic mechanisms to ensure the stability of the latent dynamics, we propose a corrective signal that builds upon \cite{kolter2019learning} and \cite{khansari2014learning}.

The contributions of this paper are: a) a Neural ODE framework for learning DS with \textit{multiple attractors} and proven stability guarantees, b) a novel loss that can capture trajectory similarity in the phase space instead of the time domain, c) formulation of the necessary components to correctly set the attractors in the latent space through diffeomorphic output mappings and d) enabling LfD with \textit{multiple feedback modalities} and without the need for \textit{state derivatives}. Overall, the aim of this paper is to introduce a flexible DS learning paradigm that requires minimal overhead, is versatile and can capture intricate motions. These contributions are made clear in Table \ref{table} where we also compare them with the characteristics of other similar methods.

\IEEEpeerreviewmaketitle

\section{Backgournd}
\label{sec:background}
\subsection{Dynamical Systems}
The task that the majority of DS learning methods consider is to learn a continuous time autonomous dynamical system given by the equation:
\begin{align}
\label{eq:dynamicalsystem}
    \dot{\boldsymbol x} = \boldsymbol f(\boldsymbol x(t)),\ t\in T
\end{align}
where $t$ is time, $T$ a time interval, $\boldsymbol x(t)\in \boldsymbol X\subset\boldsymbol R^{n_x}$ is the state of the DS and $\boldsymbol f: \boldsymbol X \longrightarrow \boldsymbol X$ a potentially nonlinear vector field to be learnt. In the particular case where this DS acts as a trajectory generator for a robotic system, $\boldsymbol x(t)$ represents the desired configuration of the robot in addition to the state of any other agent or object that interacts with it.

This formulation assumes that the state $\boldsymbol x$ of the DS is available for measurement. Furthermore, conventional DS learning approaches \cite{khansari2011learning, khansari2014learning, rana2020euclideanizing} necessitate access to state derivative information for the supervised learning of the underlying function $f$. This prerequisite presents significant challenges when endeavoring to learn dynamics within latent spaces or when derivative information is not readily accessible. To illustrate, consider a scenario where human joint trajectories, obtained from vision-based techniques \cite{toshev2014deeppose}, serve as the training data. In such cases, obtaining derivative information typically entails numerical differentiation, a process prone to yielding sub-optimal outcomes and contingent upon factors such as sampling frequency and temporal regularity.

\subsection{Neural ODEs}
In light of these considerations, we opt for Neural ODEs to model DS. This choice is motivated by their capability to address the aforementioned challenges with minimal adjustments.

\textbf{Definition 1 (Neural ODE)}: Assume an observation $\boldsymbol y\in \boldsymbol Y$ that could potentially be any sensing modality. A Neural ODE is defined as the following ODE system:
\begin{align}
\label{eq:neuralode}
\left\{
\begin{matrix*}[l]
    &\boldsymbol x(t_0) = \boldsymbol \phi_{\boldsymbol \theta}(\boldsymbol y) & (input\ mapping)\\
    &\dot{\boldsymbol x} = \boldsymbol f_{\boldsymbol \theta}(\boldsymbol x) &(latent\ state\ dynamics)\\
    &\boldsymbol z(t) = \boldsymbol \psi_{\boldsymbol \theta}(\boldsymbol x(t)) &(output\ mapping)
\end{matrix*}
\right.
\end{align}
where $\boldsymbol x\in \boldsymbol X\subset\boldsymbol R^{n_x}$ is the state variable of the ODE, $\boldsymbol z(t) \in \boldsymbol Z\subset\boldsymbol R^{n_z}$ is the output variable $\boldsymbol \phi_{\boldsymbol \theta}(\cdot):\boldsymbol R^{n_y} \longrightarrow \boldsymbol R^{n_x}$ and $\boldsymbol \psi_{\boldsymbol \theta}(\cdot):\boldsymbol R^{n_x} \longrightarrow \boldsymbol R^{n_z}$ are the input and output functions. The vector field $\boldsymbol f_{\boldsymbol \theta}: \boldsymbol R^{n_x} \longrightarrow \boldsymbol R^{n_x}$ determines the evolution of the dynamics and together with the input and output functions, it is a network with learnable parameters $\boldsymbol \theta$. The ODE is solved along an arbitrary time interval $T=[t_0,\ t_f]$ where $t_0$ and $t_f$ are the initial and final time respectively.

Neural ODEs present a framework conducive to the acquisition of deep DS operating within a latent variable $\boldsymbol x$. This inherent characteristic renders them particularly advantageous in applications within robotics and control domains where DS play a pivotal role. Despite their ability in capturing intricate behaviors and processes, Neural ODEs lack an intrinsic mechanism to ensure that the solution of the learned Ordinary Differential Equation (ODE) remains bounded, thus mitigating the risk of divergence for specific initial values within the phase space $\boldsymbol X$. In the realm of robotics, assuring stability is imperative as it correlates closely with safety and predictability.

There have been various attempts \cite{rodriguez2022lyanet, richards2018lyapunov} to address this concern; however, the majority fall short in providing stability guarantees spanning the entirety of the phase space. Typically, assurances are confined to regions encountered during training, thereby offering weaker guarantees than established DS learning methods \cite{khansari2011learning, khansari2014learning, rana2020euclideanizing}.

\section{Stable Neural ODEs for DS learning}
\label{sec:stablenodes}
In this section we present StableNODEs, which builds upon techniques that ensure stability in DS. 

\textbf{Definition 2 (StableNODE):} Given an observation $\boldsymbol y$, a StableNODE is defined by:
\begin{align}
\label{eq:stablenode}
\left\{
\begin{matrix*}[l]
    &\boldsymbol x(t_0) = \boldsymbol \phi_{\boldsymbol \theta}(\boldsymbol y) &\\
    &\dot{\boldsymbol x} = \hat{\boldsymbol f}(\boldsymbol x) &\ \ , t\in T\\
    &\boldsymbol z(t) = \boldsymbol \psi_{\boldsymbol \theta}(\boldsymbol x(t)) &
\end{matrix*}
\right.
\end{align}
where $\hat{\boldsymbol f}(\boldsymbol x) = \boldsymbol f_{\boldsymbol \theta}(\boldsymbol x) + \boldsymbol u(\boldsymbol x)$, with:
\begin{align}
\label{eq:correctiveterm}
    \boldsymbol u(\boldsymbol x) = \left\{
\begin{matrix*}[c]
0&,  L(\boldsymbol x)\leq 0 \\
    - \frac{\nabla V_{\boldsymbol \theta}(\boldsymbol x)L(\boldsymbol x) + \epsilon s L(\boldsymbol x) \boldsymbol f_{\boldsymbol \theta}(\boldsymbol x)}{|\nabla V_{\boldsymbol \theta}(\boldsymbol x)|^2 + \epsilon}&,  0<L(\boldsymbol x)<\frac{1}{s}\\
    - \frac{\nabla V_{\boldsymbol \theta}(\boldsymbol  x)L(\boldsymbol x) + \epsilon f_{\boldsymbol \theta}(\boldsymbol x)}{|\nabla V_{\boldsymbol \theta}(\boldsymbol x)|^2 + \epsilon}&,  L(\boldsymbol x)\geq\frac{1}{s}\\
\end{matrix*} \right.
\end{align}
where $L(\boldsymbol x) = \nabla V_{\boldsymbol \theta}(\boldsymbol x)\boldsymbol f_{\boldsymbol \theta} + \alpha V_{\boldsymbol \theta}(\boldsymbol x)$, $\epsilon, \alpha$ and $s$ are positive constants and $V_{\boldsymbol \theta}(\cdot)$ is a candidate Lyapunov function. The rest of the notation is consistent with Definition 1. Note that although the Lyapunov, input, output and state mappings are different networks, they are all trained jointly according to the gradients of the Neural ODE loss.   

The rationale behind function \( \hat{\boldsymbol f} \) is that it mirrors the nominal function \( \boldsymbol f_{\boldsymbol \theta} \) when \( L(\boldsymbol x) \leq 0 \). Here, the condition \( L(\boldsymbol x) \leq 0 \) signifies that the nominal dynamics exhibit exponential stability with respect to the candidate Lyapunov function \( V_{\boldsymbol \theta} \), thus requiring no alteration. However, when this condition is not met, the nominal dynamics are adjusted in a manner that ensures stability.

The formulation of function \( \hat{\boldsymbol f} \) expands upon the ideas presented in \cite{kolter2019learning, khansari2014learning}, allowing for a broader class of learnable Lyapunov functions. Notably, function \( \hat{\boldsymbol f} \) and the functions proposed in \cite{kolter2019learning} coincide when \( \epsilon = 0 \). In \cite{khansari2014learning}, the Lyapunov function is represented as a sum of asymmetric quadratic terms, while in \cite{kolter2019learning}, it takes the form of an Input Convex Neural Network (ICNN), resulting in convex Lyapunov functions characterized by a single global minimum. However, for Lyapunov stability, it is crucial that the candidate Lyapunov function be positive definite, satisfying \( V(\boldsymbol 0) = 0 \) and \( V(\boldsymbol x) > 0 \) elsewhere (where \(\boldsymbol 0\) denotes the attractor without loss of generality), which is a less strict constraint than convexity. By employing function (4), we broaden the scope to include Lyapunov functions that need not necessarily be convex. While still required to serve as candidate Lyapunov functions, they are permitted to exhibit local minima. Though the avoidance of these minima is not guaranteed, the issue of undefined dynamics in their vicinity is resolved, ensuring computational stability. Such functions prove advantageous in scenarios involving multiple attractors or where constraints on the phase space of the DS need to be imposed in the form of potential functions. Additionally, the inclusion of \( \epsilon \) in the denominator of (4) yields computational benefits in the forward solution of the ODE, sometimes mitigating stiffness problems of the latent ODE.

The stability guarantees of StableNODEs are presented in the following theorem: 

\textbf{Theorem 1:} \textit{Assume the dynamics: 
\begin{align}
\label{eq:latentdynamics}
    \dot{\boldsymbol x} = \hat{\boldsymbol f}(\boldsymbol x)
\end{align}
where $\hat{\boldsymbol f}(\boldsymbol x)$ is given by the sum of a nominal model and the corrective signal (\ref{eq:correctiveterm}). Assume a finite set of desired attractors $\boldsymbol X_e$ and a candidate Lyapunov function that satisfies $V_{\boldsymbol \theta}(\boldsymbol x)>0,\ \forall \boldsymbol x\in \boldsymbol X\setminus \{\boldsymbol X_e\}$ and $V_{\boldsymbol \theta}(\boldsymbol x) = 0,\  \forall \boldsymbol x\in \boldsymbol X_e$. Assume $\boldsymbol X_c$ is a set of any critical points of $V_{\boldsymbol \theta}$ that do not belong in $\boldsymbol X_e$. Then the solutions of (\ref{eq:latentdynamics}) converge asymptotically to the set $\boldsymbol E = \{\boldsymbol x\in \boldsymbol X| V_{\boldsymbol \theta}(\boldsymbol x)\leq\frac{1}{s\alpha}\} \cup \boldsymbol X_c$.
}

\textit{Proof:} For the proof see the Appendix

\subsection{Attractors in the output space}
The latent space of Neural ODEs often lacks a direct physical interpretation within the context of the problem they are applied to. The most significant variables, in many cases, are the output variables, as they are the ones that can be interpreted by the user. For instance, a StableNODE could be utilized to learn a DS capable of replicating the trajectory of an end effector of a manipulator. In this scenario, the latent space of the StableNODE holds no inherent physical meaning to the problem, as both functions \( \boldsymbol \phi_{\boldsymbol \theta}(\cdot) \) and \( \boldsymbol \psi_{\boldsymbol  \theta}(\cdot) \) are learned. However, the output corresponds to the positions of the robot's end effector. Consequently, a commonly acceptable choice for the attractor of the learned DS would be the final point in the demonstrated trajectory. Nevertheless, as the attractor for the StableNODE exists in the latent space, there is no direct way to set it to a desired value, and ensuring that \( \boldsymbol \psi_{\boldsymbol  \theta}(\mathbf{0}) \) coincides with that point cannot be guaranteed. Therefore, it is imperative to devise a mechanism that allows the user to define an attractor set \( \boldsymbol Z_e\) in the output space that ensures it can be directly mapped to an attractor set $\boldsymbol X_e$ in the latent space.

The set of attractors $\boldsymbol X_e$ of the latent dynamics (\ref{eq:latentdynamics}) can be modified according to the needs of the problem, by constructing an appropriate Lyapunov function that satisfies the conditions of Theorem 1. In order to specify $\boldsymbol X_e$, a continuous mapping that can map $\boldsymbol Z_e$ to $\boldsymbol X_e$ and is invertible, such that $\boldsymbol X_e$ can be mapped directly to $\boldsymbol Z_e$, is needed. Mappings $\boldsymbol \psi: \boldsymbol R^{n_x}\longrightarrow\boldsymbol R^{n_x}$ that satisfy these conditions are called bijective and if used as output mappings, they allow determining $\boldsymbol X_e$ as $\boldsymbol X_e=\{\boldsymbol \psi^{-1}(\boldsymbol z), \ \forall \boldsymbol z\in \boldsymbol Z_e\}$.

This raises the question of how to construct learnable bijective mappings that are expressive enough to capture complex behaviors. To address this challenge, we construct our output mapping as a synthesis of $n$ simpler bijective layers, denoted as $\boldsymbol \psi_{\boldsymbol  \theta} = \boldsymbol \psi_1 \circ \boldsymbol \psi_2 \circ \ldots \circ \boldsymbol \psi_n$, where each bijective layer $\boldsymbol \psi_k: \mathbb{R}^{n_x} \longrightarrow \mathbb{R}^{n_x}$ is implemented as a coupling layer \cite{dinh2016density}. Let $\boldsymbol x \in \mathbb{R}^{n_x}$ be the input to a coupling layer $\boldsymbol \psi_{k}$. The input vector is first divided into two vectors $\boldsymbol x^\alpha \in \mathbb{R}^{\lfloor \frac{n_x}{2} \rfloor}$ and $\boldsymbol x^\beta \in \mathbb{R}^{\lceil \frac{n_x}{2} \rceil}$. Subsequently, one of the two parts, with the order reversed in the subsequent layer, undergoes a scaling and translation operation. The output of layer $k$ is given by:
\begin{align}
\label{eq:couplinglayer}
\boldsymbol \psi_k(\boldsymbol x) = \begin{bmatrix*}[c]
    \boldsymbol x^\alpha \\
    \boldsymbol x^\beta\odot exp(\boldsymbol s_k(\boldsymbol x^\alpha)) + \boldsymbol t_k(\boldsymbol x^\alpha)
\end{bmatrix*}
\end{align}
where $\odot$ denotes the Hadamard product $\boldsymbol s_k:R^{\lfloor \frac{n_x}{2}\rfloor}\longrightarrow R^{\lfloor \frac{n_x}{2}\rfloor}$ and $\boldsymbol t_k:R^{\lfloor \frac{n_x}{2}\rfloor}\longrightarrow R^{\lfloor \frac{n_x}{2}\rfloor}$ are learnable scaling and translation functions respectively, implemented as feedforward neural networks. This mapping and any composition of finitely many such mappings is guaranteed to be bijective. Furthermore, the inverse is straightforward and cheap to compute. The readers are referred to  \cite{dinh2016density} for more details of this mapping and to  \cite{rana2020euclideanizing} for the stability properties of the output space dynamical system. 

\begin{figure}[t]
    \centering
    \includegraphics[width=1\linewidth]{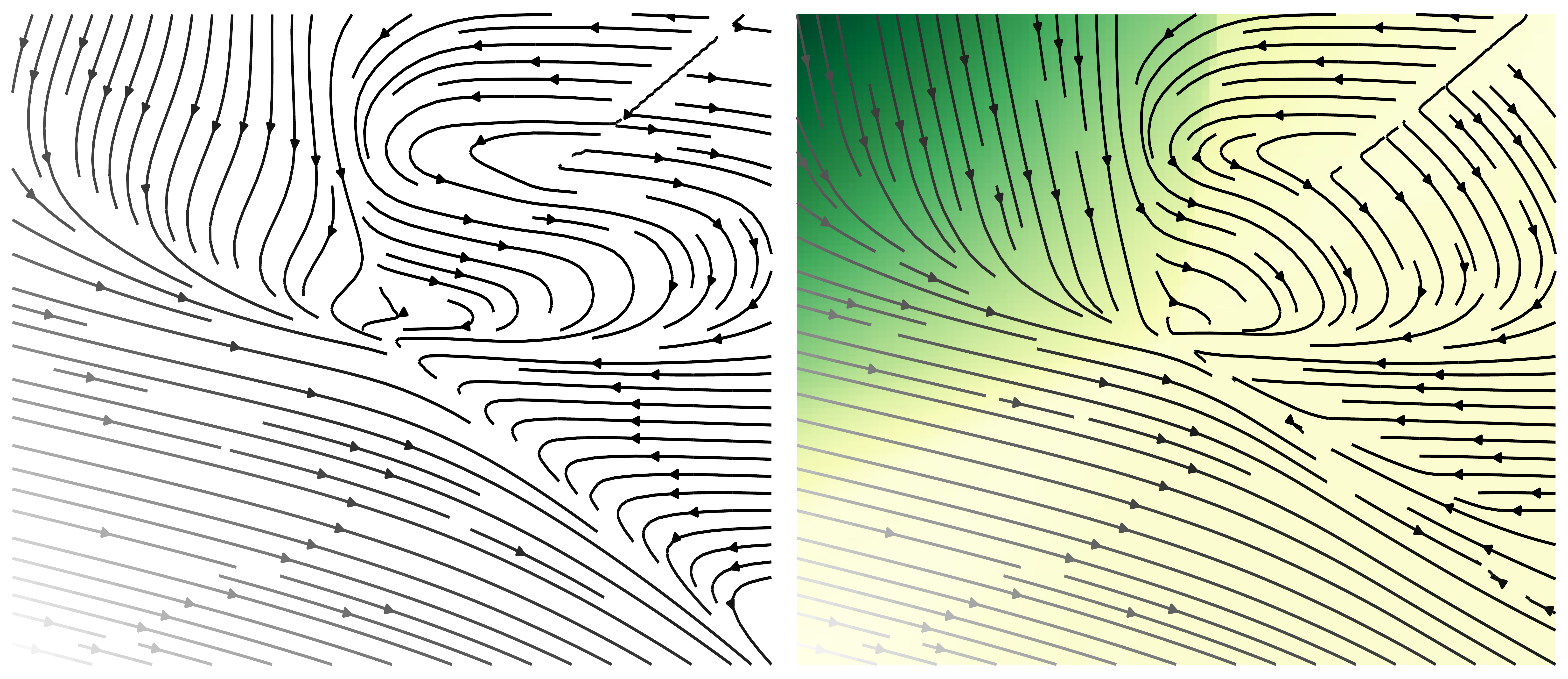}
    \caption{(a) The vector field of the nominal function $\boldsymbol f_{\boldsymbol \theta}$ for a 2 dimensional latent space and  (b) The vector field of the sum of $\boldsymbol f_{\boldsymbol \theta}$ and the corrective term $\boldsymbol u$. Dark green regions denote the areas of the phase space where $L(\boldsymbol x) > 0.$}
    \label{fig:figure2}
\end{figure}
\begin{figure}[t]
    \centering
    \includegraphics[width=1\linewidth]{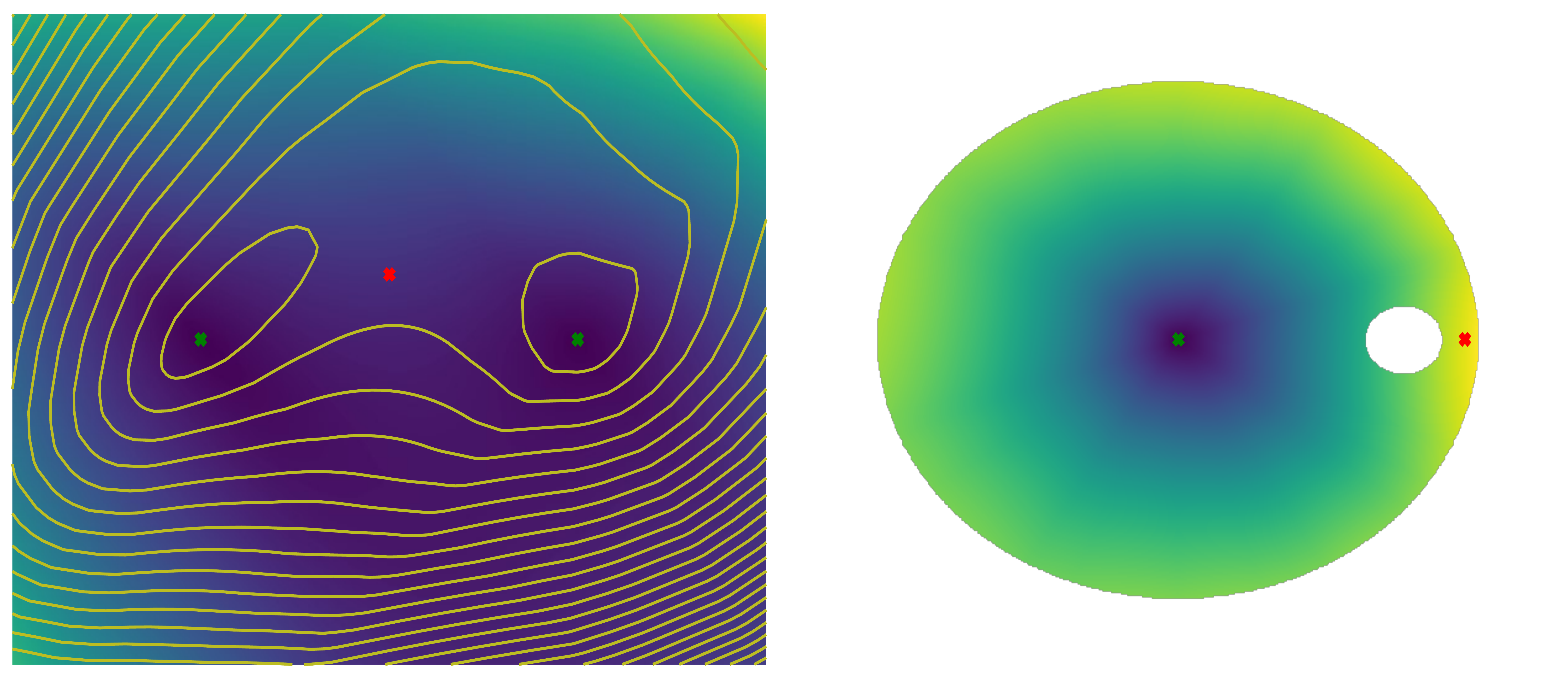}
    \caption{(a) The Lyapunov function of a 2 dimensional DS with two target points shown in green. The red point is a critical point of $V_{\boldsymbol \theta}$ and in this case is either a local maximum or a saddle point. This Lyapunov function can be useful if the DS has to learn reaching or placing motions with 2 goal positions, (b) The Lyapunov function of a DS with one target point. The Lyapunov function is build to be maximal on the boundaries of an outer circle and of an inner circle restricting thus the resulting trajectories. It is impossible to create such functions that do not exhibit other critical points than the target (target shown in green, critical point in red).}
    \label{fig:figure3}
\end{figure}
\subsection{Average Hausdorff distance as a loss function}
Neural ODEs have predominantly been employed in classification problems \cite{rodriguez2022lyanet, chen2018neural}, where the ODE is solved forward in time, yielding the final time value as the model's prediction. Consequently, the loss function becomes a function of \( \boldsymbol x(t_f) \), independent of intermediate trajectory values. However, when Neural ODEs are utilized to model DS, such as in robot planning and control or human motion prediction, the loss function typically encompasses the entire trajectory \( \boldsymbol x_{t_0:t_f} \). This distinction impacts the selection of the loss function and the approach to computing its gradients with respect to the trainable parameters. The former will be addressed in this subsection, while the latter will be discussed in the subsequent section.

\begin{figure}[t]
    \centering
    \includegraphics[width=1\linewidth]{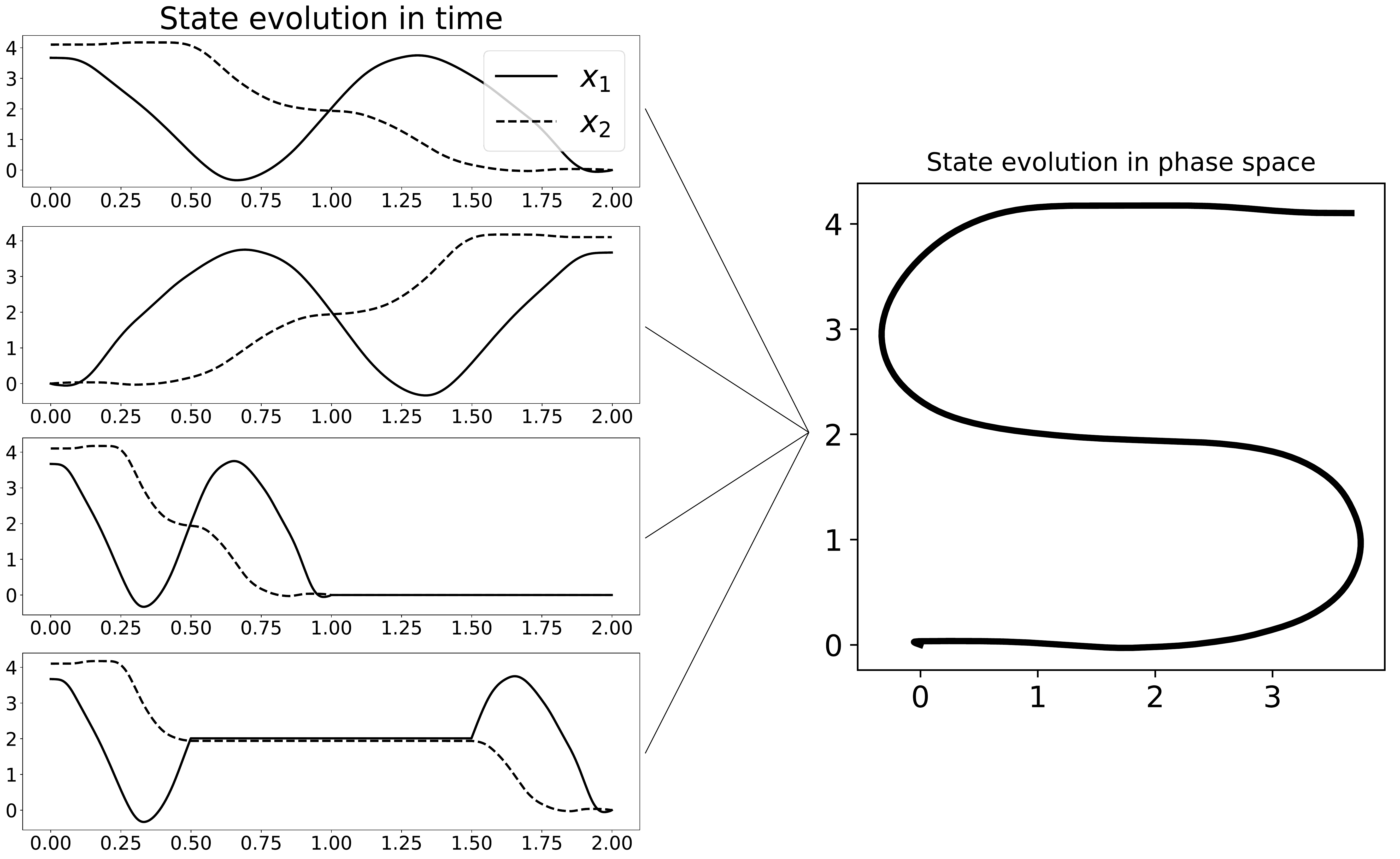}
    \caption{Four trajectories with time misalignments that correspond to the same phase space trajectory.}
    \label{fig:figure4}
\end{figure}

Demonstrations used to train robots in acquiring skills are typically untimed, and tasks are often non-time-critical. The primary objective is for the robot to reproduce the demonstrated trajectory, regardless of whether it executes it slightly faster or slower. However, a Mean Squared Error (MSE) loss, comparing the simulated and actual trajectories, not only measures trajectory similarity in phase space but also captures timing discrepancies. Consequently, even if a trajectory is identical to the demonstrated one in phase space, it could incur a high MSE loss due to timing differences. This phenomenon is visible in Fig. \ref{fig:figure4}. Moreover, besides deviating from our objective, an MSE loss imposes strict training criteria, necessitating more training iterations and increased network parameters.

Furthermore, considering \( \hat{\boldsymbol f} \), there's a significant hyperparameter to be determined, with substantial implications for the learning process and its outcome. The parameter \( \alpha \) dictates the rate of Lyapunov function decay and is selected a priori. However, since the Lyapunov function is unknown beforehand, choosing \( \alpha \) is not straightforward. Large values can render the optimization problem challenging to solve, as it may be improbable to find a Lyapunov function that decreases exponentially with rate \( \alpha \) for a \( \hat{\boldsymbol f} \) perfectly tracking desired trajectories. Nevertheless, by minimizing a trajectory similarity loss in the phase space, we afford \( \hat{\boldsymbol f} \) and \( V_{\boldsymbol \theta} \) more flexibility to be learned while adhering to the imposed constraints. For these reasons, we opt to employ the Average Hausdorff Distance (AHD) \cite{schutze2012using} as the loss function between predicted and actual trajectories.

Consider two unordered non-empty sets $\hat{\boldsymbol X}$ and $\boldsymbol X_d$ and a distance metric $d(\boldsymbol x,\boldsymbol x_d)$ between two points $\boldsymbol x\in \hat{\boldsymbol X}$ and $\boldsymbol x_d\in \boldsymbol X_d$. The distance metric could be any metric, however in our case the Euclidean distance in the phase space is used. Note that the two sets do not have to be of the same size. For problems addressed in this paper, $\hat{\boldsymbol X}$ is a trajectory generated by solving the StableNODE forward in time, while $\boldsymbol X_d$ is a demonstrated trajectory. Then the AHD is defined as:
\begin{align}
\label{eq:hausdorrfloss}
    L_{H} = \frac{1}{|\hat{\boldsymbol X}|}\sum_{\boldsymbol x\in \hat{\boldsymbol X}}\min_{\boldsymbol x_d\in \boldsymbol X_d}d(\boldsymbol x,\boldsymbol x_d) + \frac{1}{|\boldsymbol X_d|}\sum_{\boldsymbol x_d\in \boldsymbol X_d}\min_{\boldsymbol x\in \hat{\boldsymbol X}}d(\boldsymbol x,\boldsymbol x_d)
\end{align}
Loss (\ref{eq:hausdorrfloss}), if implemented within an autograd framework, like PyTorch, is differentiable for all $(\boldsymbol x,\boldsymbol x_d)$ although it is not strictly continuous. Notice how the definition does not require the two sets $\hat{\boldsymbol X}$ and $\boldsymbol X_d$ to be ordered. Trajectories can be considered as sets ordered by time, however as mentioned in the previous paragraphs, the trace of the trajectories on the phase plane is of importance. Moreover, the two sets don't have to be of the same size in order to calculate the loss, which makes it easier to calculate using the solution of an adaptive solver and irregularly sampled demonstrations.

\subsection{Computing loss gradients}
Training a Neural ODE is equivalent to solving an optimization problem of the form:

\begin{align}\label{eq:neuralodeloss}
    &\min_\theta \ L[\boldsymbol x(t, \boldsymbol \theta)]  \\
    s.t. \ \ \ \ \  & \dot{\boldsymbol x} = \hat{\boldsymbol f}(\boldsymbol x) \nonumber \\
    & \boldsymbol x(t_0) = \boldsymbol \phi_{\boldsymbol  \theta}(\boldsymbol y) \nonumber
\end{align}

with respect to \( \boldsymbol \theta \), where \( L[\boldsymbol x(t, \boldsymbol \theta)] = \int_{t_0}^{t_f} l(\boldsymbol x(t,\boldsymbol \theta)) \, dt + l_f(\boldsymbol x(t_f,\boldsymbol \theta)) \) represents a loss functional. Here, \( l \) and \( l_f \) denote arbitrary instantaneous and final costs, respectively. These losses can take the form of similarity metrics, such as (\ref{eq:hausdorrfloss}), comparing the Neural ODE-generated trajectory with demonstrated trajectories, or encode any other desirable properties. If the demonstrations include state derivatives, they could be utilized as labels for training \( \hat{\boldsymbol f}\) in a regression fashion. However, the optimization problem (\ref{eq:neuralodeloss}) does not assume availability of state derivatives and allows for more intricate costs to be minimized through the functions $l$ and $l_f$. Subsequently, solving (\ref{eq:hausdorrfloss}) is challenging, as it requires computing gradients throughout the solution of the latent ODE.

The derivatives of the functional (\ref{eq:neuralodeloss}) can be computed using the \textit{state adjoint method} \cite{kidger2022neural}, which involves solving the latent ODE forward in time and the adjoint ODE backward in time \cite{chen2018neural}. This approach offers significant benefits in terms of memory cost for calculating derivatives compared to using ODE solvers implemented within an auto-differentiation framework. However, solving two ODEs, as required by the \textit{state adjoint method}, can be considerably slower when evaluating the functional (\ref{eq:neuralodeloss}) over a large time period.

For computational efficiency reasons, auto-differentiation through the ODE solver is used as the default gradient calculation method for (\ref{eq:neuralodeloss}) throughout this paper, unless otherwise mentioned.

\section{Experiments}
\subsection{Results on the LASA dataset}
The effectiveness of StableNODEs in learning demonstrated trajectories is evaluated using the benchmark LASA dataset \cite{khansari2011learning}. This open-source dataset, comprises 30 handwritten shapes, with seven different demonstrations provided for each shape, resulting in a total of 210 trajectories. The performance of Euclideanizing Flows (EF)\cite{rana2020euclideanizing} and Deep Stable Dynamics (DSD) \cite{kolter2019learning} is also assessed and compared against StableNODEs. 

To capture the expressiveness of the latent space formulation utilized in StableNODEs, we adopt the identity maps for both input and output mappings. The nominal function \( \boldsymbol f_{\boldsymbol \theta} \) is modeled using a feedforward network, while the Lyapunov function is represented by an ICNN. We employ the training loss (\ref{eq:hausdorrfloss}) and obtain solutions of the ODEs using the \textit{Dopri5} solver implemented in PyTorch with the package \textit{torchdiffeq} \cite{chen2018neural}. The parameters for the corrective term are set as follows: \( \epsilon = 1 \times 10^{-5} \), \( \alpha = 1 \times 10^{-3} \), and \( s = 20 \). For DSD, we utilize the same nominal model as for StableNODE, set \( \epsilon \) to 0, use the Mean Squared Error (MSE) loss, and obtain solutions using the \textit{Euler} solver. EFs are implemented following the method described in \cite{rana2020euclideanizing}. 

All three methods are evaluated using the Dynamic Time Warping Distance (DTWD) and the Discrete Frechet Distance (FD), both of which are time-invariant measures of time-series similarity. These metrics effectively capture the similarity between the demonstrated and recreated trajectories in the phase space and are commonly employed to evaluate DS learning methods \cite{rana2020euclideanizing, zhang2022learning}.

\begin{figure}[t]
    \centering
    \includegraphics[width=1\linewidth]{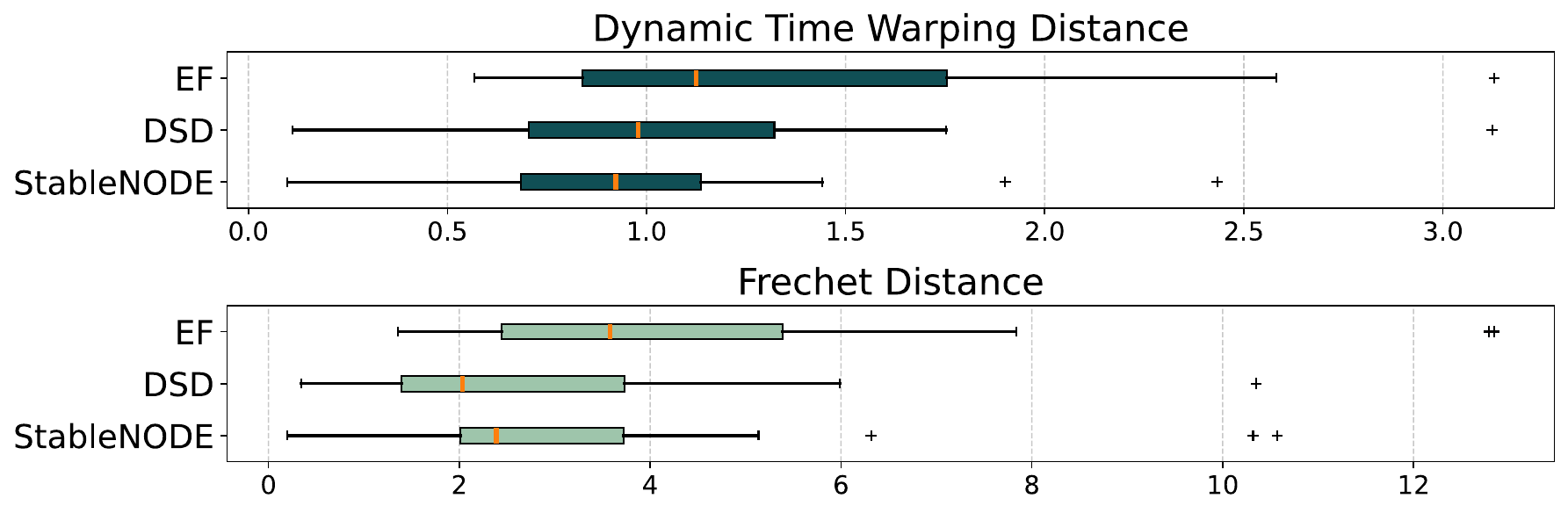}
    \caption{Dynamic Time Warping Distance and Frechet Distance of Euclideanizing Flows (EF), Deep Stable Dynamics (DSD) and StableNODEs.}
    \label{fig:figure5}
\end{figure}

As depicted in Fig. \ref{fig:figure5}, StableNODE consistently outperforms the other two DS learning methods across multiple shapes. StableNODE demonstrates a lower median and standard deviation of DTWD scores throughout the dataset, while DSD exhibits a lower median Frechet score. Notably, both StableNODE and DSD outperform EF, which could potentially be attributed to the fact that velocity information required for EF's training was obtained through finite differentiation. It is also worth mentioning that data resampling and smoothing operations were necessary to achieve the performance with EF, whereas raw trajectories were used for learning with DSD and StableNODE.

Although DSD and StableNODE demonstrate similar performance, it is conceivable that StableNODEs would outperform DSD in scenarios with smaller sampling frequencies, irregularly spaced data, or a mixture of fast and slow modes in the trajectories, owing to the utilization of adaptive solvers. 

\begin{figure}[t]
    \centering
    \includegraphics[width=1\linewidth]{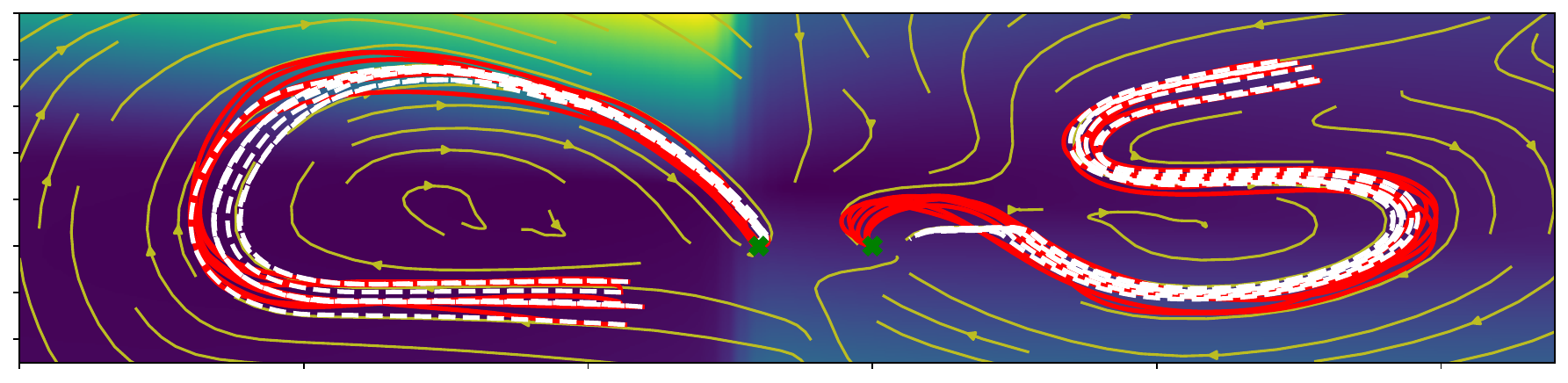}
    \caption{Demonstrated trajectories (solid red line) and DS generated trajectories (white dashed line) of a DS with two target points.}
    \label{fig:figure6}
\end{figure}

The capability of StableNODE to model DS with multiple attractors is also assessed by combining the translated demonstrations from two different shapes, as illustrated in Fig. \ref{fig:figure6}. To capture this behavior, a Lyapunov function of the form \( V_{\boldsymbol \theta}(\boldsymbol x) = \sigma_\gamma(\boldsymbol x)V_1(\boldsymbol x - \boldsymbol x^*_{1}) + (1- \sigma_\gamma(\boldsymbol x))V_2(\boldsymbol x - \boldsymbol x^*_{2}) \) is employed, where \( V_1 \) and \( V_2 \) are represented by ICNNs, \( \boldsymbol x^*_{1} = [0,\ 0]^T \) and \( \boldsymbol x^*_{2} = [0,\ -0.2]^T \) denote the two attractors, and \( \sigma_\gamma(\boldsymbol x) \) represents the sigmoid function with a positive scale factor \( \gamma = 70 \).

From Fig. \ref{fig:figure6}, it is evident that StableNODE can generate trajectories from either shape that converge to the correct attractor. However, performance is slightly compromised due to the overlapping of the two Lyapunov functions; this could potentially be minimized by placing the attractors further apart.

\begin{figure}[t]
    \centering
    \includegraphics[width=1\linewidth]{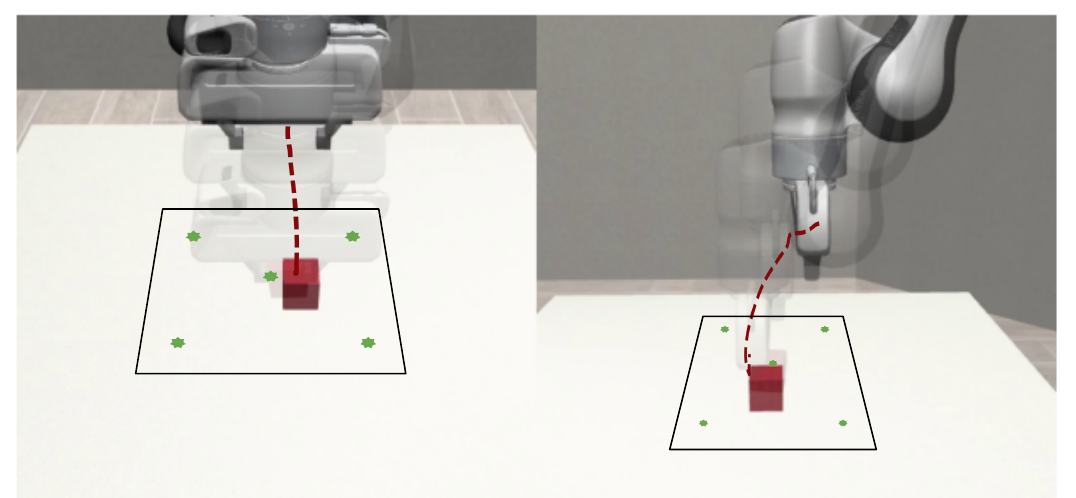}
    \caption{Generated end-effector trajectory using a RGB image. The green stars represent the 5 attractors of the DS and the black box denotes the surface within which the cube can be spawned.}
    \label{fig:figure7}
\end{figure}
\begin{figure*}[t]
    \centering
    \includegraphics[width=1\linewidth]{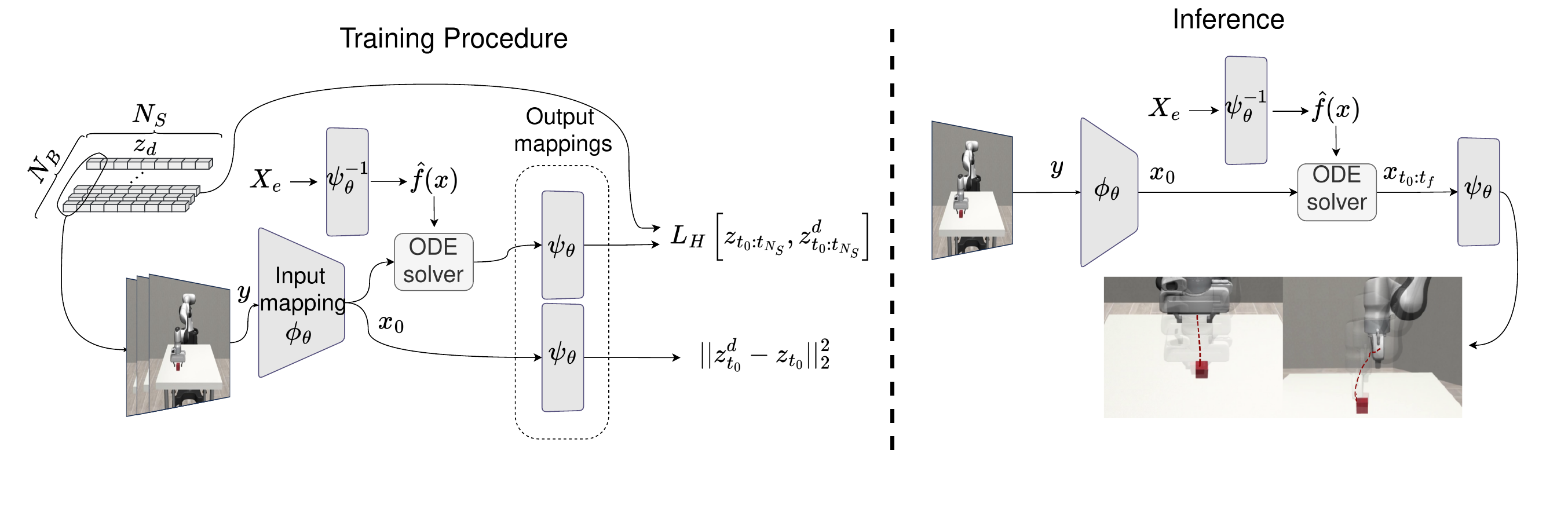}
    \caption{The end-to-end training process for a StableNODE to generate trajectories from a single RGB image (left) and the trajectory generation process (right). $N_B$ denotes the number of training batches and $N_S$ the size of the training trajectories.}
    \label{fig:figure8}
\end{figure*}
\subsection{Image-to-trajectory generation for a simulated picking task}
The efficacy of StableNODEs as generative models for image-to-trajectory generation is evaluated in a simulated picking task, involving a Franka Emika  robot and a red cube placed on a table. This simulation environment is part of Robosuite \cite{robosuite2020} and has been used in other works as a benchmark for evaluating vision-based policy learning techniques \cite{robomimic2021}. We tackle a similar problem, that of generating a trajectory for the end-effector of the robot using a single RGB image of the initial state of the workspace, as can be seen in Fig. \ref{fig:figure8}. The position of the cube is considered unknown both during training and inference and it is sampled uniformly within a 0.03 by 0.02 rectangle at the beginning of each trial. 

Demonstrations of successful executions of the task from proficient humans were used for training a StableNODE. These demonstrations were gathered using the RoboTurk \cite{mandlekar2018roboturk} teleoperation platform and they are part of the datasets provided within the Robomimic framework \cite{robomimic2021}. 120 of such demonstrations were used for the training of the StableNODE.

The StableNODE used consists of a 4 layer convolutional input network and an output mapping of three stacked coupling layers given by (\ref{eq:couplinglayer}). As depicted in Fig. \ref{fig:figure8}, during a training a step, a batch of $N_B$ demonstrations of length $N_S$ is randomly sampled. The RGB images $\boldsymbol y\in\boldsymbol R^{84\times 84}$ collected at the initial time sample of the batched trajectories, are used as inputs to the input network $\boldsymbol \phi_{\boldsymbol  \theta}(\cdot)$, which in turn produces a point $\boldsymbol x_0$ in the latent space of the StableNODE. This point is then used as an initial value to an ODE solver according to eq. (\ref{eq:neuralode}). The solution of the ODE and $\boldsymbol x_0$ concurrently pass through the output mapping to produce a generated trajectory $\boldsymbol z_{t_0:N_S}$ and an image-to-robot position prediction $\boldsymbol z_{t_0}$. The training loss is then calculated as:
\begin{align}
\label{eq:exploss}
    L = L_H\left[ \boldsymbol z_{t_0:t_{N_S}}, \boldsymbol z^d_{t_0:t_{N_S}} \right] + k\cdot||\boldsymbol z_{t_0} - \boldsymbol z^d_{t_0}||^2
\end{align}
We found that the second term of eq. (\ref{eq:exploss}) helps align the initial position of the generated trajectory with that of the end effector, since end effector position information is not assumed to be available. The latent dynamics $\hat{\boldsymbol f}$, together with a Lyapunov function $V_{\boldsymbol \theta}$, implemented as an ICNN, are learnt by minimizing $L_H$. The hyperparameters used in our experiments are $k = 15,\ N_B= 120, \ N_S = 25, \ \epsilon = 10^{-4}, \ s=10$ and $\alpha = 1$. The output $z\in\boldsymbol R^6$ of the StableNODE is the end-effector position and velocity. 

Since the cube position is unknown, it cannot be used as an attractor, which means that grasping the cube is part of the transient behavior of the DS. We hypothesize that introducing attractors located near the edges and center of the rectangular area the cube lies in, the DS can be steered to any location the cube is in. In order to achieve this, we introduce 5 attractors: $\boldsymbol z^*_0 = [0, 0, 0.82, 0, 0, 0]^T$ and $\boldsymbol z_i^* =[\pm 0.025, \pm0.015, 0.82, 0, 0, 0]^T, i=1,2,3,4$, as seen in Fig. \ref{fig:figure7}, and 5 ICNNs $V_i,\ i=0,1,...,5$ associated with each attractor. A Lyapunov function is constructed as ${\displaystyle V = \prod_{i=0}^4 V_i(\boldsymbol x - \boldsymbol \psi_{\boldsymbol  \theta}^{-1}(\boldsymbol z^*_i))}$ .

During inference an image is fed to the trained StableNODE, which in turn produces a trajectory for the end-effector. We evaluated the ability of the generated trajectories to effectively solve the task of picking the cube over 1000 trials, with the position of the cube sampled uniformly from a 0.03 by 0.02 rectangle at every trial. The task was considered complete once the cube was lifted from the table. Furthermore, the control of the end-effector is considered to be out of scope for this problem and is thus implemented using the robot's and the cube's position. 

Our method achieved a success percentage of $91.27\%$, despite the tracking errors introduced by the robot controller. In the failed attempts, although the robot failed to grasp the cube, its behaviour was stable and it always approached one of the attractors. Furthermore, a StableNODE trained on only 6 demonstrations was evaluated in order to assess the generalization capabilities of the method in the absence of data for a large part of the phase space. The method was successful $73.53\%$ of the trials, exhibiting good behaviour for cube locations close to the ones seen in training, but failing when the cube is placed close to the boundaries of the rectangular sampling area.

\section{Conclusions}
\label{sec:conclusions}
In this study, we introduce a framework utilizing NeuralODEs for the learning of stable DS. Our approach augments Neural ODEs by incorporating a corrective function, ensuring provable stability across the entire phase space, even in the presence of multiple attractors. Additionally, we propose an output mapping mechanism comprising learnable diffeomorphisms, facilitating the definition and transfer of attractors between the output and latent spaces. We employ the Average Hausdorff Distance (AHD) as the training loss, as it offers a time-invariant metric for evaluating trajectory similarity. To validate the effectiveness of our contributions, we conduct comparative analyses against existing methods using a publicly available dataset. Moreover, we train StableNODEs to generate trajectories from RGB images for a picking task, achieving a success rate of $91.27\%$. Overall, StableNODEs present a modular framework adaptable to various problem domains by adjusting the training loss, attractor specifications, or the type of output mappings. Potential future expansions of our approach include mitigating unwanted critical points in Lyapunov functions and extending the capability to learn discontinuous DS for generating trajectories with contacts \cite{bianchini2023simultaneous}.

\section*{Appendix}
\label{sec:appendix}
\subsection*{A. Proof of Theorem 1}
 We split the Lyapunov analysis in three distinct cases:

$\bullet$ If $ L(\boldsymbol x) \leq 0 $ function $V_{\boldsymbol \theta}$ exponentially decreases since:
\begin{align*}
    \nabla V_{\boldsymbol \theta}(\boldsymbol x)\hat{\boldsymbol f}(\boldsymbol x) &\leq -\alpha V_{\boldsymbol \theta}(\boldsymbol x) \implies \\
    \dot{V_{\boldsymbol \theta}}(\boldsymbol x) &\leq -\alpha V_{\boldsymbol \theta}(\boldsymbol x) \implies \\
     V_{\boldsymbol \theta}(\boldsymbol x) &\leq |V_{\boldsymbol \theta}(0)|e^{-\alpha t}
\end{align*} 

$\bullet$ If $ 0 < L(\boldsymbol x) < \frac{1}{s}$ then the derivative of the Lyapunov function is :
\begin{align*}
\dot{V_{\boldsymbol \theta}}(\boldsymbol x) &=  \nabla V_{\boldsymbol \theta}^T(\boldsymbol x)\hat{\boldsymbol f}(\boldsymbol x) \implies\\
\dot{V_{\boldsymbol \theta}}(\boldsymbol x) &= \left(1 - \frac{\epsilon s L(\boldsymbol x)}{|\nabla V_{\boldsymbol \theta}(\boldsymbol x)|^2 + \epsilon}\right)\nabla V^T_\theta(\boldsymbol x) \boldsymbol f_{\boldsymbol \theta}(\boldsymbol x) \\
&\ \ \ \ \ \ \ \ \ \ \ - \left(1 - \frac{\epsilon }{|\nabla V_{\boldsymbol \theta}(\boldsymbol  x)|^2 + \epsilon}\right)L(\boldsymbol x)
\end{align*}
After expanding $L(\boldsymbol x) =  V^T_\theta(\boldsymbol x) \boldsymbol f_{\boldsymbol \theta}(\boldsymbol x) + \alpha V_{\boldsymbol \theta}(\boldsymbol  x)$ and simplifying:
\begin{align*}
    \dot{V_{\boldsymbol \theta}}(\boldsymbol  x) &= \frac{\epsilon - \epsilon s L(\boldsymbol  x)}{|\nabla V_{\boldsymbol \theta}(\boldsymbol  x)|^2 + \epsilon}\nabla V^T_\theta(\boldsymbol  x) \boldsymbol f_{\boldsymbol \theta}(\boldsymbol  x)\\ &\ \ \ \ \ \ \ \ - \frac{|\nabla V_{\boldsymbol \theta}(\boldsymbol  x)|^2 }{|\nabla V_{\boldsymbol \theta}(\boldsymbol  x)|^2 + \epsilon}\alpha V_{\boldsymbol \theta}(\boldsymbol  x)
\end{align*}
Since $- \frac{|\nabla V_{\boldsymbol \theta}(\boldsymbol  x)|^2 }{|\nabla V_{\boldsymbol \theta}(\boldsymbol  x)|^2 + \epsilon}\alpha V_{\boldsymbol \theta}(\boldsymbol  x) \leq 0$:
\begin{align*}
    \dot{V_{\boldsymbol \theta}}(\boldsymbol  x) \leq \frac{\epsilon - \epsilon s L(\boldsymbol  x)}{|\nabla V_{\boldsymbol \theta}(\boldsymbol  x)|^2 + \epsilon}\nabla V^T_\theta(\boldsymbol  x) \boldsymbol f_{\boldsymbol \theta}(\boldsymbol  x)
\end{align*}
Moreover, owing to the fact that $L(\boldsymbol  x)< \frac{1}{s}$ it holds that $1-sL(\boldsymbol  x) >0$ and $\nabla V^T_\theta(\boldsymbol  x) \boldsymbol f_{\boldsymbol \theta}(\boldsymbol  x) < \frac{1}{s} - \alpha V_{\boldsymbol \theta}(\boldsymbol  x)$. After substituting these expressions in the inequality for $\dot{V_{\boldsymbol \theta}}(\boldsymbol  x)$, we get:
\begin{align*}
\dot{V_{\boldsymbol \theta}}(\boldsymbol  x) &< \frac{\epsilon - \epsilon s L(\boldsymbol  x)}{|\nabla V_{\boldsymbol \theta}(\boldsymbol  x)|^2 + \epsilon} \left(\frac{1}{s} - \alpha V_{\boldsymbol \theta}(\boldsymbol  x)\right)     
\end{align*}
The derivative of the candidate Lyapunov function is rendered negative when $V_{\boldsymbol \theta}(\boldsymbol  x) > \frac{1}{s\alpha}$. This implies that in the regions of the phase space where $0< L(\boldsymbol  x)< \frac{1}{s}$, $V_{\boldsymbol \theta}$ is decreasing until it reaches the set $\{\boldsymbol x\in X| V_{\boldsymbol \theta}(\boldsymbol  x)\leq\frac{1}{s\alpha}\}$. Although the level sets of $V_{\boldsymbol \theta}$ after the training are not known a priori, the set $\{\boldsymbol x\in X| V_{\boldsymbol \theta}(\boldsymbol  x)\leq\frac{1}{s\alpha}\}$ can be made arbitrary small by increasing $s$.
 
$\bullet$ If $ L(\boldsymbol  x) \geq \frac{1}{s} $ then the derivative of the candidate Lyapunov function is:
\begin{align*}
    \dot{V_{\boldsymbol \theta}}(\boldsymbol  x) &=  \left(1  -\frac{\epsilon }{|\nabla V_{\boldsymbol \theta}(\boldsymbol  x)|^2 + \epsilon}\right)\nabla V_{\boldsymbol \theta}^T(\boldsymbol  x)f_{\theta}(\boldsymbol  x) \\
    &\ \ \ \ \ \ \ \ - |\nabla V_{\boldsymbol \theta}(\boldsymbol  x)|^2 \frac{\dot{V_{\boldsymbol \theta}}(\boldsymbol  x)}{|\nabla V_{\boldsymbol \theta}(\boldsymbol  x)|^2 + \epsilon} \implies \\
    \dot{V_{\boldsymbol \theta}}(\boldsymbol  x) &= -\alpha \left(1  -\frac{\epsilon }{|\nabla V_{\boldsymbol \theta}(\boldsymbol  x)|^2 + \epsilon}\right) V_{\boldsymbol \theta}(\boldsymbol  x) 
\end{align*}
If $V_{\boldsymbol \theta}$ is a convex function with respect to $x$ then if $L(\boldsymbol  x) \geq \frac{1}{s} \implies \left(1  -\frac{\epsilon }{|\nabla V_{\boldsymbol \theta}(\boldsymbol  x)|^2 + \epsilon}\right)> 0$, since $|\nabla V_{\boldsymbol \theta}(\boldsymbol  x)|^2 > 0$. That means that $V_{\boldsymbol \theta}$ is decreasing exponentially, except in the regions where $0<L(\boldsymbol  x)<\frac{1}{s}$. If $\epsilon = 0$ then the same analysis as \cite{kolter2019learning} is followed. 

If $\boldsymbol X$ is a closed and bounded set, then if $V_{\boldsymbol \theta}$ is made maximal on the boundary $\partial \boldsymbol X$, then it is guaranteed that $\boldsymbol X$ is forward invariant if $\boldsymbol x_0 \in \boldsymbol X$. This is because $V_{\boldsymbol \theta}$ is decreasing which means that $\boldsymbol x$ never reaches $\partial \boldsymbol X$.

If $\boldsymbol X_c$ is not empty then the strictest form the last equality can get is $\dot{V}_\theta(\boldsymbol  x) = 0$, which would imply that the DS will be trapped in $\boldsymbol X_c$. Eventually, $\boldsymbol x$ will be asymptotically confined within the set $\boldsymbol E$. Note that avoiding critical points would require either special classes of Lyapunov functions \cite{koditschek1990robot} or a discontinuous DS \cite{4518905}. Although entrapment in critical points is not optimal, our method accounts for it by avoiding the explosion of $\hat{\boldsymbol f}$ to infinity when they are encountered. Note that although function $\hat{\boldsymbol f}$ is not continuously differentiable, $V_{\boldsymbol \theta}$ is continuously differentiable since $\hat{\boldsymbol f}$ is continuous.


\printbibliography

\end{document}